\documentclass[11pt,a4paper]{article}
\usepackage[hyperref]{acl2021}
\usepackage{times}
\usepackage{latexsym}
\usepackage{xspace}
\usepackage{graphicx}
\usepackage{subfig}
\usepackage{float}

\usepackage{microtype}
\usepackage{booktabs, makecell, siunitx}
\usepackage{multirow}
\usepackage{colortbl}

\aclfinalcopy 

\title{Itih\={a}sa: A large-scale corpus for Sanskrit to English translation}
  
\author{\textbf{Rahul Aralikatte}$^{1}$ \textbf{Miryam de Lhoneux}$^{1}$ \textbf{Anoop Kunchukuttan}$^{2}$ \textbf{Anders Søgaard}$^{1}$ \\   
    $^1$University of Copenhagen~~~$^2$Microsoft AI and Research \\
    $^1${\tt \{rahul,ml,soegaard\}@di.ku.dk} \\
    $^2${\tt ankunchu@microsoft.com} \\
}

\newcommand\itihasa{Itih\={a}sa\xspace}
\newcommand\ramayana{The R\={a}m\={a}yana\xspace}
\newcommand\mahabharata{The Mah\={a}bh\={a}rata\xspace}
\definecolor{Gray}{gray}{0.92}

\begin{document}
\maketitle

\begin{abstract}
This work introduces \itihasa, a large-scale translation dataset containing 93,000 pairs of Sanskrit {\it shloka}s and their English translations. The {\it shloka}s are extracted from two Indian epics viz., \ramayana and \mahabharata. We first describe the motivation behind the curation of such a dataset and follow up with empirical analysis to bring out its nuances. We then benchmark the performance of standard translation models on this corpus and show that even state-of-the-art transformer architectures perform poorly, emphasizing the complexity of the dataset.\footnote{The processed and split dataset can be found at \url{https://github.com/rahular/itihasa} and a human-readable version can be found at \url{http://rahular.com/itihasa}.}
\end{abstract}

\section{Introduction}

Sanskrit is one of the oldest languages in the world and most Indo-European languages are influenced by it \cite{sanskit-influence}. There are about 30 million pieces of Sanskrit literature available to us today \cite{goyal-etal-30mil}, most of which have not been digitized. Among those that have been, few have been translated. The main reason for this is the lack of expertise and funding. An automatic translation system would not only aid and accelerate this process, but it would also help in democratizing the knowledge, history, and culture present in this literature. In this work, we present \itihasa, a large-scale Sanskrit-English translation corpus consisting of more than 93,000 {\it shlokas} and their translations.

\begin{figure}
    \centering
    \includegraphics[width=\columnwidth]{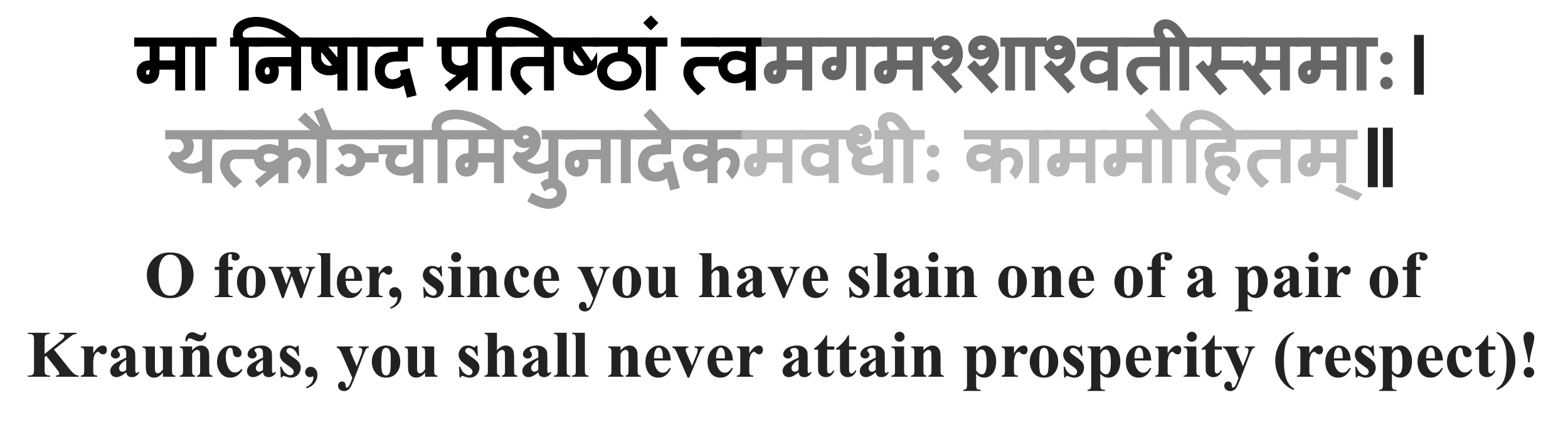}
    \caption{An introductory {\it shloka} from \ramayana. The four parts with eight syllables each are highlighted with different shades of gray.}
    \label{fig:intro}
\end{figure}

\itihasa, literally meaning `it happened this way' is a collection of historical records of important events in Indian history. These bodies of work are mostly composed in the form of verses or {\it shloka}s, a poetic form which usually consists of four parts containing eight syllables each (Fig. \ref{fig:intro}). The most important among these works are \ramayana and \mahabharata. \ramayana, which describes the events in the life of Lord R\={a}ma, consists of 24,000 verses. \mahabharata details the war between cousins of the Kuru dynasty, in 100,000  verses. \mahabharata is the longest poem ever written with about 1.8 million words in total and is roughly ten times the length of the Iliad and the Odyssey combined.

\noindent Only two authors have attempted to translate the unabridged versions of both \ramayana and \mahabharata to English: Manmatha N\={a}th Dutt in the 1890s and Bibek Debroy in the 2010s. M. N. Dutt was a prolific translator whose works are now in the public domain. These works are published in a {\it shloka}-wise format as shown in Fig. \ref{fig:intro} which makes it easy to automatically align {\it shloka}s with their translations. Though many of M. N. Dutt's works are freely available, we choose to extract data from \ramayana \cite{ramayana}, and \mahabharata \cite{mahabharata}, mainly due to its size and popularity. As per our knowledge, this is the biggest Sanskrit-English translation dataset to be released in the public domain.

\begin{figure*}
    \centering
    \includegraphics[width=\textwidth]{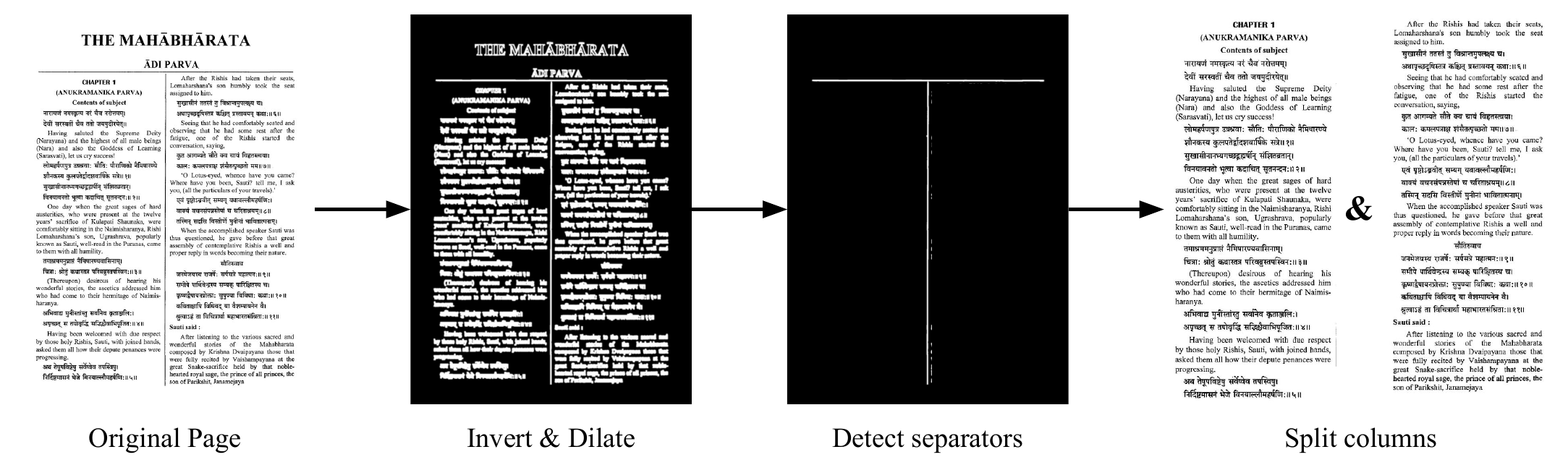}
    \caption{Pre-processing Pipeline: The three steps shown here are: (i) invert the colour scheme of the PDF and dilate every detectable edge, (ii) find the indices of the longest vertical and horizontal lines in the page, and (iii) split the original PDF along the found separator lines.}
    \label{fig:preproc}
\end{figure*}

\noindent We also train and evaluate standard translation systems on this dataset. In both translation directions, we use Moses as an SMT baseline, and Transformer-based seq2seq models as NMT baselines (see \S\ref{sec:expts}). We find that models which are generally on-par with human performance on other translation tasks, perform poorly on \itihasa, with the best models scoring between 7-8 BLEU points. This indicates the complex nature of the dataset (see \S\ref{sec:analysis} for a detailed analysis of the dataset and its vocabulary).

\paragraph{Motivation}

The main motivation behind this work is to provide an impetus for the Indic NLP community to build better translation systems for Sanskrit. Additionally, since \ramayana and \mahabharata are so pervasive in Indian culture, and have been translated to all major Indian languages, there is a possibility of creating an {\it n}-way parallel corpus with Sanskrit as the pivot language, similar to Europarl \cite{europarl} and PMIndia \cite{pmindia} datasets.

The existence of Sanskrit-English parallel data has other advantages as well. Due to Sanskrit being a morphologically rich, agglutinative, and highly inflexive, complex concepts can be expressed in compact forms by combining individual words through {\it Sandhi} and {\it Samasa}.\footnote{{\it Sandhi} refers to the concatenation of words, where the edge characters combine to form a new one. {\it Samasa} can be thought of as being similar to elliptic constructions in English where certain phrases are elided since their meaning is obvious from the context.} This also enables a speaker to potentially create an infinite number of unique words in Sanskrit. Having a parallel corpus can help us induce word translations through bilingual dictionary induction \cite{bdi}. It also allows us to use English as a surrogate language for tasks like knowledge base population. Constituency or dependency parsing, NER, and word sense disambiguation can be improved using indirect supervision \cite{indirect-supervision}. Essentially, a parallel corpus allows us to apply a plethora of transfer learning techniques to improve NLP tools for Sanskrit.

\section{Data Preparation}

The translated works of \ramayana and \mahabharata were published in four and nine volumes respectively.\footnote{The digitized (scanned) PDF versions of these books are available at \url{https://hinduscriptures.in}} All volumes have a standard two-column format as shown in Fig. \ref{fig:preproc}. Each page has a header with the chapter name and page number separated from the main text by a horizontal line. The two columns of text are separated by a vertical line. The process of data preparation can be divided into (i) automatic OCR extraction, and (ii) manual inspection for alignment errors.

\paragraph{Automatic Extraction}

\begin{figure*}[htb!]
\centering
\subfloat[Print error.\label{fig:print_err}]{\includegraphics[width=0.33\textwidth]{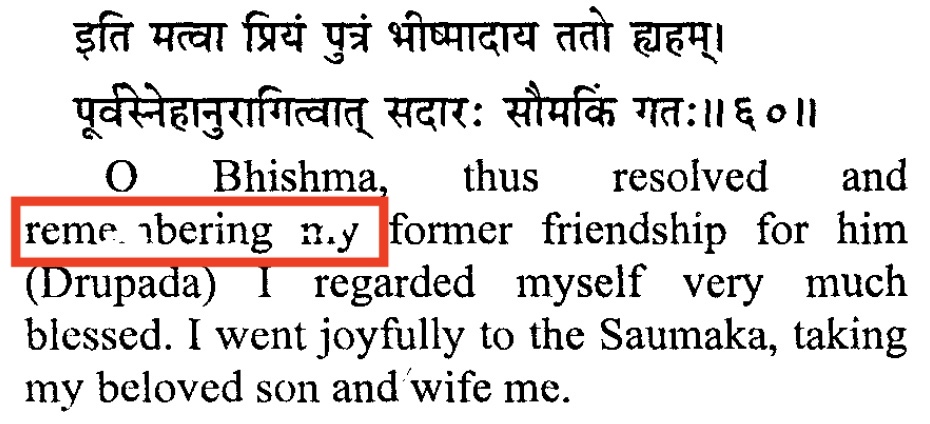}}\hfill
\subfloat[Input error.\label{fig:input_err}] {\includegraphics[width=0.33\textwidth]{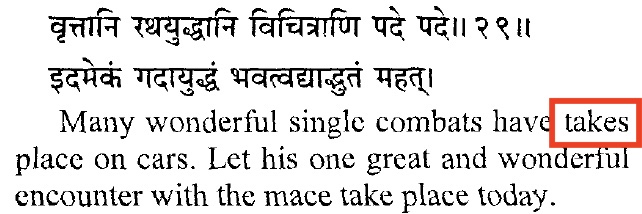}}\hfill
\subfloat[Subjective error.\label{fig:sub_err}] {\includegraphics[width=0.33\textwidth]{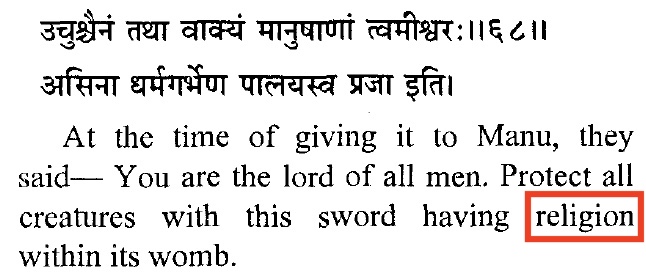}}\hfill
\caption{Different types of errors found in the original text while performing manual inspection.} \label{fig:errors}
\end{figure*}

The OCR systems we experimented with performed poorly on digitized documents due to their two-column format. They often fail to recognize line breaks which result in the concatenation of text present in different columns. To mitigate this issue, we use an edge detector\footnote{We invert the color scheme and apply a small dilation for better edge detection using OpenCV \cite{opencv}.} to find the largest horizontal and vertical lines, and using the indices of the detected lines, split the original page horizontally and vertically to remove the header and separate the columns (see Fig. \ref{fig:preproc}). We then input the single-column documents to Google Cloud's OCR API\footnote{More information can be found at \url{https://cloud.google.com/vision/docs/pdf}} to extract text from them. To verify the accuracy of the extracted text, one chapter from each volume (13 chapters in total) is manually checked for mistakes. We find that the extracted text is more than 99\% and 97\% accurate in Sanskrit and English respectively. The surprising accuracy of Devanagari OCR can be attributed to the distinctness of its alphabet. For English, this number decreases as the OCR system often misclassifies similar-looking characters (viz., {\it e} and {\it c}, {\it i} and {\it l}, etc.).

\paragraph{Manual Inspection}

An important limitation of the OCR system is its misclassification of alignment spaces and line breaks. It sometimes wrongly treats large gaps between words as line breaks and the rest of the text on the line is moved to the end of the paragraph which results in translations being misaligned with its {\it shloka}s. Therefore, the output of all 13 volumes was manually inspected and such misalignments were corrected.\footnote{This was a time-consuming process and the first author inspected the output manually over the course of one year.}

Upon manual inspection, other kinds of errors were discovered and corrected where possible.\footnote{It was not feasible for the authors to correct every error, especially the lexical ones. The most common error that exists in the corpus is the swapping of {\it e} and {\it c}. For example, `thcir' instead of `their'. Though these errors can easily be corrected using automated tools like the one proposed in \cite{error-correct}, it is out-of-scope of this paper and is left for future work.} These errors can be categorized as follows: 
(i) {\it print errors}: this type of error is caused by occluded or faded text, smudged ink, etc. An example can be seen in Fig. \ref{fig:print_err}, 
(ii) {\it input errors}: these are human errors during typesetting the volumes which include typos (Fig, \ref{fig:input_err}), exclusion of words, inclusion of spurious words, etc., 
(iii) {\it subjective errors}: these are contextual errors in the translation itself. For example, in Fig. \ref{fig:sub_err}, the word {\it dharma} is incorrectly translated as `religion' instead of `righteousness', and
(iv) {\it OCR errors}: these errors arise from the underlying OCR system. An example of such errors is the improper handling of split words across lines in the Devanagari script. If the OCR system encounters a hyphen as the last character of a line, the entire line is ignored. In general, print errors are corrected as much as possible, subjective errors are retained for originality, and other types of errors are corrected when encountered.

\begin{table}
    \centering
    \small
    \begin{tabular}{r|ccc|c}
    \toprule
    & Train & Dev. & Test & Total \\
    \midrule
    \rowcolor{Gray}\multicolumn{5}{c}{Rāmayana} \\
    Chapters & 514 & 42 & 86 & 642 \\
    Shlokas & 15,834 & 1,115 & 2,422 & 19,371 \\
    \rowcolor{Gray}\multicolumn{5}{c}{Mahābhārata} \\
    Chapters & 1,688 & 139 & 283 & 2,110\\
    Shlokas & 59,327 & 5,033 & 9,299 & 73,659 \\
    \rowcolor{Gray}\multicolumn{5}{c}{Overall} \\
    Chapters & 2202 & 181 & 369 & 2,752 \\
    Shlokas & 75,161 & 6,148 & 11,721 & 93,030 \\
    \bottomrule
    \end{tabular}
    \caption{Size of training, development, and test sets.}
    \label{tbl:sizes}
\end{table}

\section{Analysis}\label{sec:analysis}
In total, we extract 19,371 translation pairs from 642 chapters of \ramayana and 73,659 translation pairs from 2,110 chapters of \mahabharata. It should be noted that these numbers do not correspond to the number of {\it shloka}s because, in the original volumes, {\it shloka}s are sometimes split and often combined to make the English translations flow better. We reserve 80\% of the data from each text for training MT systems and use the rest for evaluation. From the evaluation set, 33\% is used for development and 67\% for testing. The absolute sizes of the split data are shown in Tab. \ref{tbl:sizes}.

Due to Sanskrit's agglutinative nature, the dataset is asymmetric in the sense that, the number of words required to convey the same information, is less in Sanskrit when compared with English. \ramayana's English translations, on average, have 2.54 words for every word in its {\it shloka}. This value is even larger in \mahabharata with 2.82 translated words per {\it shloka} word.

\begin{figure}[t!]
    \centering
    \includegraphics[width=\columnwidth]{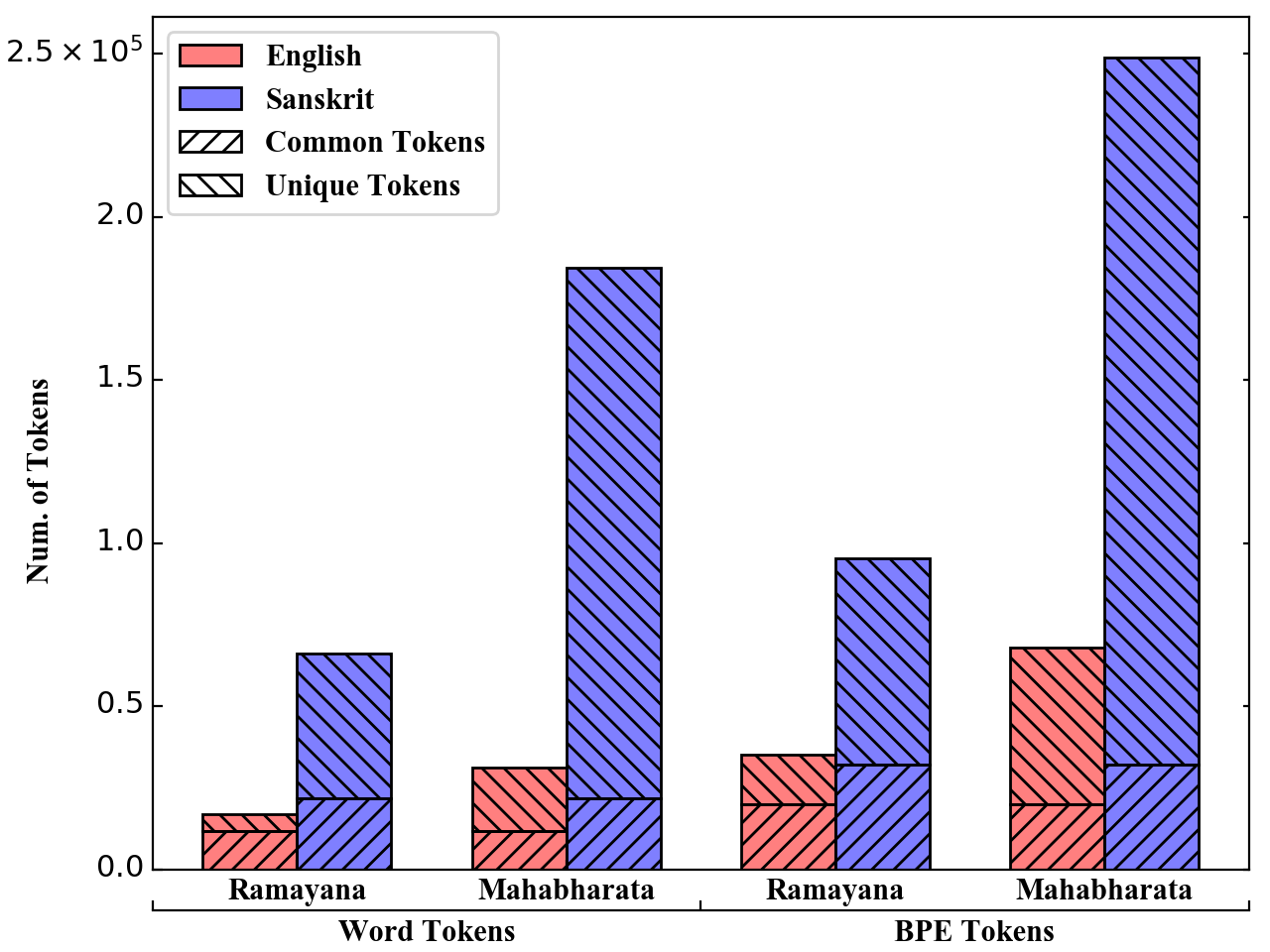}
    \caption{Comparison of vocabulary sizes. Sanskrit's morphological and agglutinative nature accounts for the large number of unique tokens in the vocabularies. }
    \label{fig:vocab}
\end{figure}

This effect is clearly seen when we consider the vocabulary sizes and the percentage of common tokens between the texts. For this, we tokenize the data with two different tokenization schemes: word-level and byte-pair encoding \cite[BPE]{bpe}. For word-level tokenization, the translations of \ramayana (\mahabharata) have 16,820 (31,055) unique word tokens, and the {\it shloka}s have 66,072 (184,407) tokens. The English vocabularies have 11,579 common tokens which is 68.8\% of \ramayana's and 37.3\% of \mahabharata's. But the overlap percentages drop significantly for the Sanskrit vocabularies. In this case, we find 21,635 common tokens which amount to an overlap of 32.7\% and 11.7\% respectively. As shown in Fig. \ref{fig:vocab}, this trend holds for BPE tokenization as well.

\section{Experiments}\label{sec:expts}
We train one SMT and five NMT systems in both directions and report the (i) character $n$-gram F-score, (ii) token accuracy, (iii) BLEU \cite{bleu}, and (iv) Translation Edit Ratio \cite[TER]{ter} scores in Tab. \ref{tbl:results}. For SMT, we use Moses \cite{moses} and for NMT, we use sequence-to-sequence (seq2seq) Transformers \cite{transformer}. We train the seq2seq models from scratch by initializing the encoders and decoders with standard BERT (B2B) architectures. These {\tt Tiny}, {\tt Mini}, {\tt Small}, {\tt Medium}, and {\tt Base} models have 2/128, 4/256, 4/512, 8/512, and 12/768 layers/dimensions respectively. See \newcite{turc2019} for more details. In our early experiments, we also tried initializing the encoders and decoders with weights from pre-trained Indic language models like MuRIL \cite{muril}, but they showed poor performance and thus are not reported here.

\paragraph{Implementation Details}
All models are trained using HuggingFace Transformers \cite{hf}. Both source and target sequences are truncated at 128 tokens. We train WordPiece tokenizers on our dataset and use them for all models. Adam optimizer \cite{adam} with weight-decay of 0.01, and learning rate of $5\times10^{-5}$ is used. All models are trained for 100 epochs. The learning rate is warmed up over 8,000 steps and decayed later with a linear scheduler. We use a batch size of 128, and use standard cross-entropy loss with no label smoothing. We run into memory errors on bigger models ({\tt medium} and {\tt base}), but maintain the effective batch-size and optimization steps by introducing gradient accumulation and increasing the number of epochs, respectively. Also, to reduce the total training time of bigger models, we stop training if the BLEU score does not improve over 10 epochs. During generation, we use a beam size of 5 and compute all metrics against truncated references.

\begin{table}
    \centering
    \small
    \begin{tabular}{r|cccc}
        \toprule
        Model & chrF & \thead{Tok. \\Acc.} & BLEU & \thead{TER\\($\downarrow$)} \\
        \midrule
        \rowcolor{Gray}\multicolumn{5}{c}{English to Sanskrit} \\
        Moses            & 25.9 & 5.48 & 0.21 & 1.53 \\ 
        B2B-{\tt Tiny}   & 15.1 & 8.07 & 2.85 & 1.04 \\
        B2B-{\tt Mini}   & 21.6 & 8.52 & 5.66 & 1.03 \\
        B2B-{\tt Small}  & 23.4 & 8.75 & 6.93 & 1.03  \\
        B2B-{\tt Medium} & 23.7 & 8.67 & 6.94 & \textbf{1.02} \\
        B2B-{\tt Base}   & \textbf{24.3} & \textbf{8.89} & \textbf{7.59} & 1.04 \\
        \rowcolor{Gray}\multicolumn{5}{c}{Sanskrit to English} \\
        Moses            & 29.3 & 8.03 & 5.67 & \textbf{0.91} \\ 
        B2B-{\tt Tiny}   & 24.5 & \textbf{8.61} & 5.64 & 0.98 \\
        B2B-{\tt Mini}   & 29.3 & 8.58 & 7.28 & 0.95 \\
        B2B-{\tt Small}  & 30.1 & 8.55 & \textbf{7.49} & 0.95 \\
        B2B-{\tt Medium} & 30.4 & 8.49 & 7.48 & 0.94 \\
        B2B-{\tt Base}   & \textbf{30.5} & 8.38 & 7.09 & 0.93 \\
        \bottomrule
    \end{tabular}
    \caption{Character F1, Token accuracy, BLEU, and TER scores for Moses and Transformer models. Scores marked with ($\downarrow$) are better if they are lower.}
    \label{tbl:results}
\end{table}

\paragraph{Discussion} We see that all models perform poorly, with low token accuracy and high TER. While the English to Sanskrit (E2S) models get better with size, this pattern is not clearly seen in Sanskrit to English (S2E) models. Surprisingly for S2E models, the token accuracy progressively decreases as their size increases. Also, Moses has the best TER among S2E models which suggests that the seq2seq models have not been able to learn even simple co-occurrences between source and target tokens. This leads us to hypothesize that the Sanskrit encoders produce sub-optimal representations. One way to improve them would be to add a {\it Sandhi-splitting} step to the tokenization pipeline, thereby decreasing the Sanskrit vocabulary size. Another natural extension to improve the quality of representations would be to initialize the encoders with a pre-trained language model.

\noindent Though it is clear that there is a large scope for improvement, the models are able to learn some interesting features of the dataset. Fig. \ref{fig:errors} shows a random gold translation pair and the \texttt{small} model's prediction. Though we see repetitions of phrases and semantic errors, the prediction follows the meter in which the original {\it shloka}s are written, i.e. it also consists of 4 parts containing 8 syllables each.

\section{Related Work}
Early translation efforts from Sanskrit to English were limited to the construction of dictionaries by Western Indologists \cite{muller, williams}. Over the years, though notable translation works like \newcite{ganguli} have been published, the lack of digitization has been a bottleneck hindering any meaningful progress towards automatic translation systems. This has changed recently, at least for monolingual data, with the curation of digital libraries like GRETIL\footnote{\url{http://gretil.sub.uni-goettingen.de/gretil.html}} and DCS\footnote{\url{http://www.sanskrit-linguistics.org/dcs/index.php}}. Currently, the largest freely available repository of translations are for The Bhagavadgita \cite{gita-iitk} and \ramayana \cite{valmiki-iitk}.

However, labeled datasets for other tasks, like the ones proposed in \cite{treebank, sandhikosh, dcs-normal} have resulted in parsers \cite{morph-parser, struct-pred} and sandhi splitters \cite{sandhi2, sandhi1} which are pre-cursors to modular translation systems. Though there have been attempts at building Sanskrit translation tools \cite{anusaaraka}, they are mostly rule-based and rely on manual intervention. We hope that the availability of the \itihasa corpus pushes the domain towards end-to-end systems.

\begin{figure}
    \centering
    \includegraphics[width=\columnwidth]{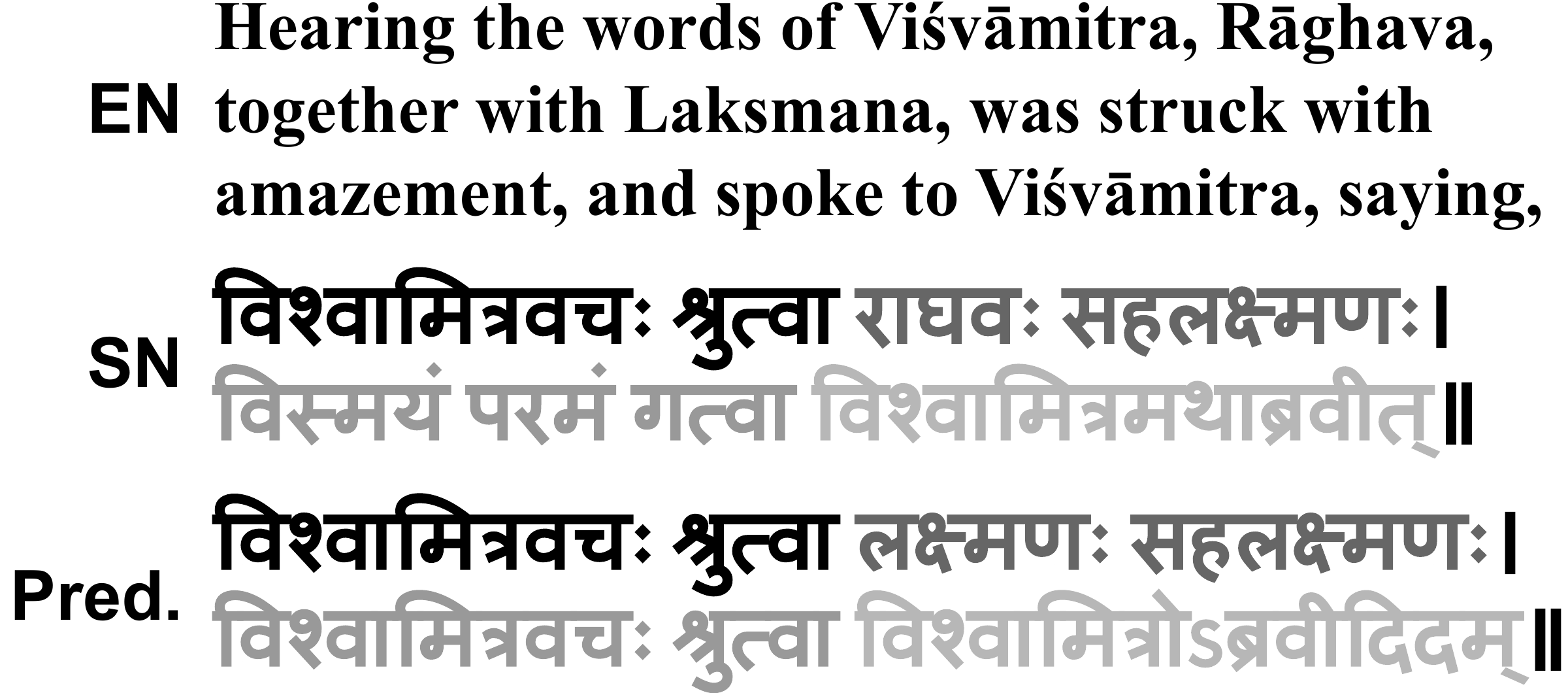}
    \caption{A gold sentence and {\it shloka} from the test set, and its corresponding \texttt{small} model prediction.}
    \label{fig:errors}
\end{figure}

\section{Conclusion}
In this work, we introduce \itihasa, a large-scale dataset containing more than 93,000 pairs of Sanskrit {\it shlokas} and their English translations from \ramayana and \mahabharata. First, we detail the extraction process which includes an automated OCR phase and a manual alignment phase. Next, we analyze the dataset to give an intuition of its asymmetric nature and to showcase its complexities. Lastly, we train state-of-the-art translation models which perform poorly, proving the necessity for more work in this area.

\section*{Acknowledgements}
We thank the reviewers for their valuable feedback. Rahul Aralikatte and Anders S{\o}gaard are funded by
a Google Focused Research Award.

\bibliography{ref}
\bibliographystyle{acl_natbib}

\end{document}